\documentclass[pmlr,twocolumn,10pt]{jmlr} 

\usepackage{booktabs}
\usepackage{siunitx}

\theorembodyfont{\upshape}
\theoremheaderfont{\scshape}
\theorempostheader{:}
\theoremsep{\newline}

\def\independenT#1#2{\mathrel{\rlap{$#1#2$}\mkern2mu{#1#2}}}
\usepackage{graphicx}
\usepackage{multirow}
\usepackage{amssymb}
\usepackage{pifont}
\usepackage{amsmath}
\usepackage{diagbox}
\usepackage{url}
\usepackage{hyperref}
\usepackage{longtable}
\newcommand{\independent}{\protect\mathpalette{\protect\independenT}{\perp}}
\usepackage{amsmath}
\usepackage{wrapfig}
\usepackage{algorithm,algcompatible,amsmath}
\algnewcommand\INPUT{\item[\textbf{Input:}]}%
\algnewcommand\OUTPUT{\item[\textbf{Output:}]}%
\jmlrvolume{LEAVE UNSET}
\jmlryear{2022}
\jmlrsubmitted{LEAVE UNSET}
\jmlrpublished{LEAVE UNSET}
\jmlrworkshop{Conference on Health, Inference, and Learning (CHIL) 2022}

\title{Multi-Task Adversarial Learning for Treatment Effect Estimation in Basket Trials}

\author{%
\Name{Zhixuan Chu} \Email{chuzhixuan.czx@alibaba-inc.com}\\
\addr Ant Group, China
\AND
\Name{Stephen L. Rathbun} \Email{rathbun@uga.edu}\\
\addr University of Georgia, USA
\AND
\Name{Sheng Li} \Email{sheng.li@uga.edu}\\
\addr University of Georgia, USA
}

\begin{document}
\maketitle
\begin{abstract}

Estimating treatment effects from observational data provides insights about causality guiding many real-world applications such as different clinical study designs, which are the formulations of trials, experiments, and observational studies in medical, clinical, and other types of research. In this paper, we describe causal inference for application in a novel clinical design called basket trial that tests how well a new drug works in patients who have different types of cancer that all have the same mutation. We propose a multi-task adversarial learning (MTAL) method, which incorporates feature selection multi-task representation learning and adversarial learning to estimate potential outcomes across different tumor types for patients sharing the same genetic mutation but having different tumor types. In our paper, the basket trial is employed as an intuitive example to present this new causal inference setting. This new causal inference setting includes, but is not limited to basket trials. This setting has the same challenges as the traditional causal inference problem, i.e., missing counterfactual outcomes under different subgroups and treatment selection bias due to confounders. We present the practical advantages of our MTAL method for the analysis of synthetic basket trial data and evaluate the proposed estimator on two benchmarks, IHDP and News. The results demonstrate the superiority of our MTAL method over the competing state-of-the-art methods.

\end{abstract}

\paragraph*{Data and Code Availability}
This paper uses benchmarks \textit{IHDP} ~\citep{brooks1992ihdp} and \textit{News} ~\citep{schwab2018perfect}, which are available on the repositories \footnote{\url{https://github.com/vdorie/npci}}
\footnote{\url{https://archive.ics.uci.edu/ml/datasets/bag+of+words}}. We also use one synthetic basket trial dataset and the detailed simulation procedure is provided in the Section~\ref{simulated_data}. 

\section{Introduction}
With the rapid development of next-generation sequencing and comprehensive genomic profiling, genomic characterization informs the treatment of a variety of cancers. Some genetic mutations have been linked to multiple cancer types; for example, BRCA1 and BRCA2 are associated with an increased risk of breast, ovarian and pancreatic cancers ~\citep{mersch2015cancers}. Traditional clinical trials focusing on patients with a single cancer are time-consuming and expensive, and frequently fail, so they are not sufficient for the development of genomic technologies. Patients are generally classified by their primary cancer and randomized controlled trials are conducted to create standard therapies for each cancer type. It is unrealistic to conduct separate clinical trials for each sub-population based on molecular subtypes or detailed classification of tumors  ~\citep{hirakawa2018master}. Therefore, a new-style clinical trial protocol is in urgent demand in oncology. 

\begin{figure}
    \begin{center}
    \includegraphics[width=0.8\linewidth]{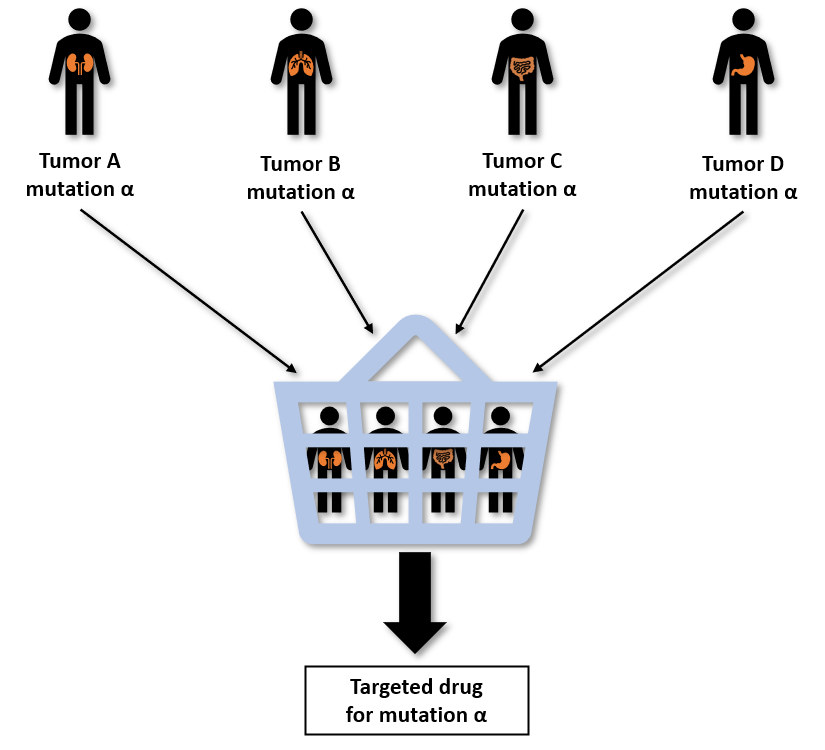}
    \end{center}
    \caption{A basket trial is usually a non-randomized and single-arm trial so that all patients with the specified genomic mutation receive the same treatment, regardless of tumor types .}
    \label{fig: Basket}
\end{figure}
A novel clinical design called basket trial has been developed based on the presence of a specific genomic mutation, irrespective of histology ~\citep{astsaturov2017future,simon2017critical,tao2018basket}. Unlike traditional clinical trials which test a drug against a specific cancer, the core organizing principle of basket trials is a common genomic mutation. A basket trial is usually a non-randomized, single-arm trial so that all patients with the specified genomic mutation receive the same treatment. Treatment selection only depends on genomic mutation type, instead of tumor type. An example of a basket trial is shown in Fig.~\ref{fig: Basket}, where the term ``basket'' arises from each collection of patients sharing a particular mutation and sub-studies for the same drug are conducted by tumor groups within the whole ``basket''. Patients enrolled in a basket trial are heterogeneous with respect to tumor type, histologic type, and patient characteristics, so the treatment effects are sensitive to population heterogeneity. Therefore, the absence of a control group becomes a limitation of treatment effect evaluation~\citep{hirakawa2018master}. Ignoring the heterogeneity of tumors may lead to failure to detect treatment effects and the inability to produce scientifically reliable findings~\citep{strzebonska2019umbrella}. Besides, focusing only on molecular therapy targeting a single mutation without considering the complexity of tumor biology may introduce bias. 

In this paper, we apply causal inference models to basket trials. Estimating causal effects from observational data has become an appealing research direction owing to the availability of data and low budget requirements compared with randomized controlled trials. This paper is the first to apply machine learning and causal inference to basket trials and explore the relationship between the traditional multiple treatments design and the basket trial design. In particular, we propose a multi-task adversarial learning (MTAL) method incorporating feature selection, multi-task representation learning, and adversarial learning to remove selection bias (tumor type heterogeneity) introduced by confounders. Our method generates all potential outcomes for each unit across all tumor types, regardless of heterogeneity from different tumor types, so that the sample size may be increased in basket trials for rare tumor types, increasing statistical power. We also define targeted group treatment effects to better describe treatment effects among sub-groups in a basket trial. We present the practicality and advantages of our MTAL method for synthetic basket trials, evaluate the proposed estimator on the IHDP and News datasets, and demonstrate its superiority over state-of-the-art methods.

\section{Related Work}
The landscape for oncology clinical trials is changing dramatically due to the advent of genomic characterization.  Among diverse master protocols~\citep{park2019systematic}, a basket trial evaluates the treatment effect of targeted therapy on patients with the same genomic mutation, regardless of tumor types. Bayesian hierarchical modeling has been proposed to adaptively borrow strength across cancer types to improve the statistical power of basket trials~\citep{berry2013bayesian,simon2017critical}. To avoid inflated type I errors in Bayesian hierarchical modeling, calibrated Bayesian hierarchical modeling has been proposed to evaluate the treatment effect in basket trials~\citep{chu2018bayesian}. As an alternative to Bayesian hierarchical modeling, we will apply powerful machine learning tools to basket trials. 

Embracing the rapid developments in machine learning, various treatment effect estimation methods for observational data have been proposed for causal inference\citep{cui2020causal,li2016matching,yao2021survey}. Balancing neural networks (BNNs) ~\citep{johansson2016learning} and counterfactual regression networks (CFRNET) ~\citep{shalit2017estimating} have been proposed to balance covariate distributions across treatment and control groups by regarding the problem of counterfactual inference as a domain adaptation problem. These models may be extended to any number of treatments even with continuous parameters, as described in the perfect match (PM) approach~\citep{schwab2018perfect} and DRNets ~\citep{schwab2019learning}. Li and Fu~\citep{li2017matching_nips17} regard counterfactual prediction as a classification problem and conduct matching based on balanced and nonlinear representations. \citep{chu2022learning} utilize the mutual information to learn the infomax and domain-independent representations to solve the selection bias between treatment and control groups.
 
\section{The Proposed Framework}
\subsection{Problem Statement}

\begin{figure}[th!]
    \begin{center}
    \includegraphics[width=0.8\linewidth]{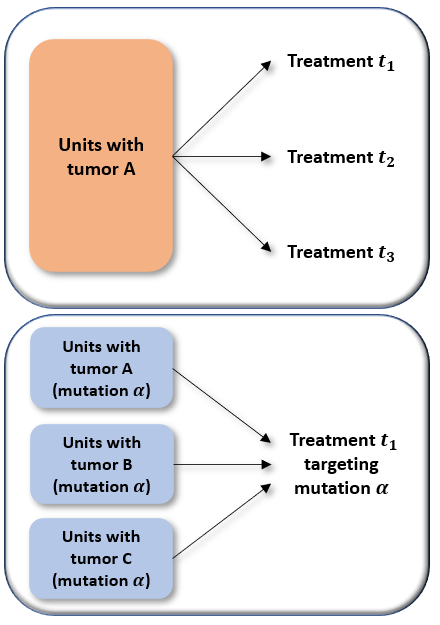}
    \end{center}
    \caption{The relationship between conventional multiple treatment causal inference (top) and basket trial (bottom).}
    \label{fig: relationship}
\end{figure}

\textbf{Clarification on New Problem Setup.} In traditional causal inference for observational data,  researchers consider binary or multiple treatments for a set of experimental units. For example, a person who has cancer may be offered a choice between two treatment therapies.  We can observe the outcome of the chosen treatment but not the potential outcome of the treatment not selected. It is impossible to observe the potential outcomes of both therapies; one of the potential outcomes is always missing. The potential outcome framework ~\citep{rubin1974estimating, splawa1990application} aims to estimate unobserved potential outcomes and then calculate the treatment effect. The basket trial tests how well a new drug works in patients who have different types of cancer with the same mutation. Patients with the same genetic mutations are put in one ``basket'' and are divided into different subgroups according to their cancer types. The differences in study design for potential outcome framework and basket trials are illustrated in Fig.~\ref{fig: relationship}. For the potential outcome framework, there is one population and several treatments, but in basket trials, there are several sub-populations and only one treatment. 

\textbf{Clarification on the Challenges.}
In the potential outcome framework ~\citep{rubin1974estimating, splawa1990application}, we mainly face two challenges: \textit{missing unobserved counterfactual outcomes} for each patient under alternative treatments not received and \textit{treatment selection bias}. In basket trials, we have the similar challenges: missing unobserved counterfactual outcomes for each patient under alternative cancers not contracted and cancer selection bias where the distributions of predictors differ among cancer types. In traditional causal inference for observational data, confounders are variables that affect both the treatment assignment and the outcome. Similarly, in basket trials, there still exist confounders that are associated with both cancer type and treatment outcome. These variables can explain why some patients with the same mutation have different types of cancer and can also influence treatment outcomes. Due to the confounders, it is difficult in a basket trial to estimate the true treatment effects of a drug targeting the mutation of interest and the true treatment effects of a drug for a specific type of cancer. If a significant treatment effect is not found, analysis of basket trials without appropriate causal inference cannot determine that the failure is due to the uselessness of the drug, the particularity of a cancer type, or individual characteristics of patients. 

\textbf{Clarification on Treatment Effects Estimation.} Because in this new setting, there is no control group, we do not care about the traditional treatment effects estimation between treated and control groups, e.g., average treatment effect (ATE) or individual treatment effect (ITE). We only focus on the counterfactual outcome inference problem, which is the core problem regardless of the new setting or traditional treatment effects estimation setting. Most basket trials are conducted as single-arm trials without a control group and a primary endpoint is given by an objective response rate (ORP). We proposed a new metric named targeted group response rate (TGOR) to better describe treatment effects in basket trials. TGOR describes the overall objective response rate for a given mutation or a given tumor type. It can evaluate the treatment effect of the drug for the whole targeted population with the same mutation and the effect of the drug in the sub-population with different specific tumors type. Our MTAL method can help remove heterogeneity across tumor types when estimating the treatment effect for targeted mutation and remove heterogeneity across patients with the same tumor when estimating the treatment effect for a targeted tumor. 

\textbf{Problem Setup.} Suppose a basket trial is conducted as a one-arm phase II trial that tests how well a new drug works in patients who have different types of tumours but share the same genetic mutation. Data are available for $n$ participants. Let $t_i \in \{1,...,k\} $ denote the type of tumour for unit $i$; $i=1,...,n$. The primary endpoint is the objective response rate (ORR)~\citep{us2007guidance}, determined by tumor assessments from radiological tests or physical examinations. Let $y_{t}^i$ denote the potential outcome of the unit $i$ ($i=1,...,n$) with the tumour $t \in \{1,...,k\}$. The observed outcome, called factual outcome is denoted by $y^f$ and the remaining unobserved potential outcomes are called counterfactual outcomes denoted by $y^{cf}$. The estimated potential, factual, and counterfactual outcomes are  $\hat{y}$, $\hat{y}^f$, and $\hat{y}^{cf}$, respectively. Let $X \in \mathbb{R}^d$ denote all observed covariates. We extend the potential outcome framework  ~\citep{rubin1974estimating} to analysis of basket trial data. The following assumptions ensure that the treatment effects can be identified: \textbf{Consistency}: The potential outcome of treatment $t$ is equal to the observed outcome if the actual treatment received is $t$. \textbf{Positivity}: For any value of $\,X$, treatment assignment is not deterministic, i.e.,$P(T = t | X = x) > 0$, for all $t$ and $x$. \textbf{Ignorability}: Given covariates $X$, treatment assignment $t$ is independent to the potential outcomes, i.e., $(y_1, y_0) \ \independent \ t | X$.

\subsection{Model Architecture}

We propose a multi-task adversarial learning (MTAL) method to analyze basket trial data or observational data in basket trials, which can remove heterogeneity across tumor types when estimating the treatment effects for a targeted mutation, remove heterogeneity among patients with one type of tumor when estimating the treatment effect for the targeted tumor, and estimating the personalized treatment effect for individual patients. Our method is also useful for studying rare cancers and cancers with rare genetic mutations by inferring the outcome of existing patients with counterfactual cancers to increase sample size and statistical power.  

\begin{figure*}[t]
    \centering
    \includegraphics[width=1\textwidth]{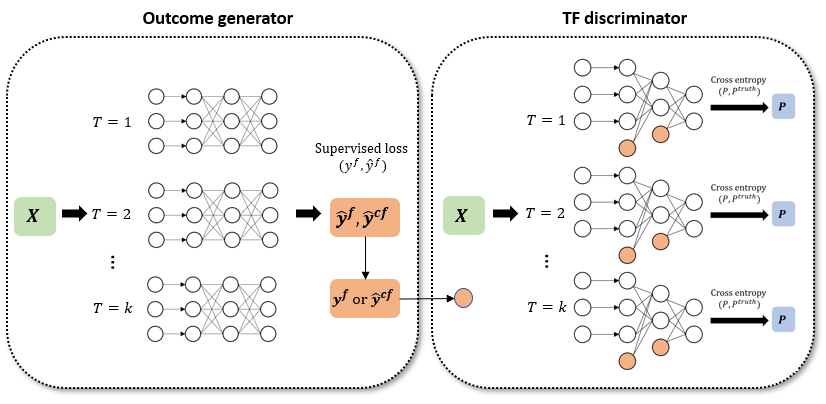}
    \vspace{-4mm}
    \caption{The framework of our multi-task adversarial learning net (MTAL). Our method contains two major components: outcome generator and true or false discriminator (TF discriminator). In the outcome generator, feature selection multi-task deep learning is utilized to estimate the potential outcomes for units across all tumor types. A deep feature selection model includes (a) a sparse one-to-one layer between the input and the first hidden layer, and (b) an elastic net regularization term throughout the fully-connected representation layers. Our TF discriminator is adopted to tell whether the outcome given the covariates and tumor type is a factual outcome. The two models are trained together in a zero-sum game and they are adversarial until the TF discriminator model is fooled by the generator.}
    \label{fig: framework}
\end{figure*}

Our method contains two major components: outcome generator and true or false discriminator (TF discriminator), as shown in Fig.~\ref{fig: framework}. In the outcome generator, we use feature selection multi-task deep learning to estimate the potential outcomes for units across all tumor types. Because different types of the tumor may have different predictor variables, which may be components of all observed covariates, a deep feature selection model including (a) a sparse one-to-one layer between the input and the first hidden layer, and (b) an elastic net regularization term throughout the fully-connected representation layers is an essential foundation for potential outcome estimation. Our TF discriminator can tell whether the outcome given the covariates and tumor type is a factual outcome. In the beginning, the TF discriminator can easily find out which outcome is a factual outcome and which one is our inferred counterfactual outcome under alternative tumor types not contracted by those patients. The outcome generator attempts to generate counterfactual outcomes in such a way that the TF discriminator cannot easily determine which is the factual outcome. These two models are trained together in a zero-sum game and they are adversarial until the TF discriminator model is fooled by the generator. At this time, we have removed the tumor type selection bias and obtained all potential outcomes for each patient across all kinds of tumors.

\subsubsection{Outcome Generator} 
Our goal is to correctly predict potential outcomes for each patient across all tumor types by a function $g: x \times t \rightarrow y$, which is parameterized by a feed-forward deep neural network structured by multiple hidden layers with non-linear activation functions. Deep neural networks can often dramatically increase prediction accuracy, describe complex relationships, and generate the structured high-level representation of features. The function $g: x \times t \rightarrow y$ uses features $x$ and tumor type $t$ as inputs to predict potential outcomes. The output of $g$ estimates potential outcomes across $k$ tumors including single factual outcome $\hat{y}^f$ and $k-1$ counterfactual outcomes $\hat{y}^{cf}$. The factual outcomes $y^f$ are used to minimize the loss of prediction $\hat{y}^f$. 

The function $g(x,t)$ maps the features and tumor type to the corresponding potential outcomes. However, when the dimension of the observed variables is high, there is a risk of losing the influence of $t$ on $g(x,t)$ if the concatenation of $x$ and $t$ is treated as input~\citep{shalit2017estimating}. To address this problem, $g(x,t)$ is partitioned into multiple head nets $g_t(x); t=\{1,...,k\}$ corresponding to each cancer type. For each cancer type, there is one independent head net for predicting the potential outcome under this tumor. Each unit is used to update only the head net corresponding to the observed tumor type. We aim to minimize the mean squared error in predicting factual outcomes by $g(x,t)$, i.e., $\mathcal{L}_Y = \frac{1}{n}\sum_{i=1}^{n}(\hat{y}_i-y_i)^2$, where $\hat{y}_i=g(x_i,t_i)$ denotes the inferred observed outcome of unit $i$ corresponding to factual treatment $t_i$. 

Due to the peculiarities of different tumor types, only a subset of all observed covariates might be predictors for each tumor type. To accommodate this, we add a deep feature selection net ~\citep{li2016deep, chu2020matching} to each head net $g_t(x), t=\{1,2,...,k\}$, which enables variable selection in deep neural networks. This model takes advantage of deep structures to capture non-linearity and conveniently selects a subset of features of the data at the input level. Feature selection at the input level can help select which variables are input into the neural network and used for representing pre-treatment variables, which makes the deep neutral network more interpretable. 

In the feature selection layer, every input variable only connects to its corresponding node where the input variable is weighted. We use a  1-1 layer instead of a fully connected layer. To select input features, weights $w$ in the feature selection layer and the following representation layers have to be sparse and only the features with nonzero weights are selected to enter the following layers. For the observational data with high dimensional variables, LASSO~\citep{tibshirani1996regression} cannot remove enough variables before it saturates. To overcome this limitation, the elastic net ~\citep{zou2005regularization} is adopted in our model, which adds a quadratic term $\lVert w \rVert_2^2$ to the penalty i.e., $\Re(w) = \lambda\lVert w \rVert_2^2+\alpha\lVert w \rVert_1$, where $\lambda$ and $\alpha$ are tuning parameters. After combining the mean squared error and the elastic net regularization term, we minimize the objective function in the outcome generator:
\begin{equation}
\begin{split}
    \mathcal{L}_g &= \frac{1}{n}\sum_{i=1}^{n}(\hat{y}_i-y_i)^2 \\
    &+\lambda \sum_{t=1}^{k}\sum_{s=1}^{S_t}\lVert w^{(s)}  \rVert_2^2 \\
    &+\alpha\sum_{t=1}^{k}\sum_{s=1}^{S_t}\lVert w^{(s)} \rVert_1,\\
    \label{Eqn1}
\end{split}
\end{equation}
where $S_t$ is the number of deep feature selection layers for the $t$-th head net including the feature selection layer and the representation layers. The $w^{(s)}$ are the parameters of deep neural network in the $s$-th layer of outcome generator. The $\lambda \geq 0$ and $ \alpha \geq 0 $ are hyperparameters that not only control the trade-off between the regularization term and the following objective terms, but also controls the trade-off between smoothness and sparsity of the weights in the feature selection layer ~\citep{li2016deep}.

\subsubsection{True or False Discriminator} 
The true or false (TF) discriminator \citep{chu2021graph} is intended to remove tumor type bias and thus improve the prediction accuracy of potential outcomes inferred in the outcome generator for each unit across all types of tumors by adversarial learning. We define one TF discriminator as  $\phi : x \times t \times (y^f \text{or} \ \hat{y}^{cf}) \rightarrow P$, where $P$ is the TF discriminator's judgement, i.e., probability that this outcome for unit $i$ given $x$ and $t$ is factual outcome, which is defined as: 
\begin{equation*}
P=  
  \begin{cases} 
   P(\text{TF judges} \ y^f \text{as factual outcome} |x,t)  
   \\ \quad\quad\quad\quad \text{if} \ t \ \text{is factual type}, \\
   P(\text{TF judges} \ \hat{y}^{cf} \text{as factual outcome} |x,t) 
   \\ \quad\quad\quad\quad \text{if} \ t \ \text{is not factual type}. \\
  \end{cases}
  \label{Eqn2}
\end{equation*}

Similar to the outcome generator, we use multiple head nets for different tumor types in the TF discriminator. In each head net, a deep feature selection net is added to select the appropriate predictors for each type of tumor. To improve the influence of $(y^f,\hat{y}^{cf})$ in the TF discriminator, we add $(y^f \text{or} \ \hat{y}^{cf})$ into each layer after one on one feature selection layer and the dimension of each layer in TF discriminator decreases layer by layer.

The TF discriminator is a binary classification task, which puts one label (i.e., true or false factual outcome) on the vector concatenating the representation vector of $x$ and potential outcome $(y^f \text{or} \ \hat{y}^{cf})$ under each type of tumor head net, so the loss of discrimination is measured by the cross-entropy with truth probability where $P^{\text{truth}}=1$ if $y^f$ is input and $P^{\text{truth}}=0$ if $\hat{y}^{cf}$ is input. In each iteration of training, we make sure to input the same number of units in each tumor type to ensure that there exist factual units in each head net. When there are several types of tumors, we face the imbalanced classification issue. If there are $k$ types of tumor and $n$ units in each tumor type are input into the model training procedure, then in each head net,  $n$ units are factual outcomes and $(k-1)n$ units are inferred as counterfactual outcomes. As $k$ increases, the imbalance of factual outcome numbers and inferred counterfactual outcome numbers in each head net will aggravate. Because inputs of TF discriminator are generated by the outcome generator $g(x,t)$, the weighted cross entropy of TF discriminator and elastic net are defined as:
\begin{equation}
\begin{split}
    \mathcal{L}_{\phi,g} = &-\frac{1}{n\times k}\sum_{t=1}^{k}\sum_{i=1}^{n}(w_0p^{\text{truth}}_{ti}\log(p_{ti})\\
    &+w_1(1-p^{\text{truth}}_{ti})\log(1-p_{ti}))\\
    &+\lambda \sum_{t=1}^{k}\sum_{r=1}^{r_t}\lVert w^{(r)} \rVert_2^2+\alpha\sum_{t=1}^{k}\sum_{r=1}^{r_t}\lVert w^{(r)} \rVert_1,
    \label{Eqn3}
\end{split}
\end{equation}
where $p^{\text{truth}}_{ti}$ is the probability that this input outcome for unit $i$ under tumor $t$ is the observed factual outcome or inferred outcome from generator module, i.e., 1 or 0, separately. $P_{ti}$ is the probability judged by TF discriminator that this input outcome for unit $i$ under tumor $t$ is factual outcome. The $w_1$ and $w_0$ are the proportions of factual outcomes and counterfactual outcomes in total outcomes. Because during training, the same number of units in each tumor type are input,  $w_1=\frac{1}{k}$ and $w_0=\frac{k-1}{k}$ in each head net. The number of deep feature selection layers for the $t$-th head net is denoted by $r_t$, and $w^{(r)}$ are the weights for the deep neural network in the $r$-th layer of the TF discriminator. $\lambda \geq 0$ and $ \alpha \geq 0 $ are the same as those in the outcome generator.

\subsubsection{Adversarial Learning} 

We have described an outcome generator to estimate potential outcomes for each unit across all types of tumor and a TF discriminator to determine if each potential outcome given unit's features under different tumor types is factual. In the initial iterations of the training algorithm, the outcome generator may generate potential outcomes that are very different from factual outcomes as determined by the TF discriminator. As the model is trained further, the TF discriminator may no longer be able to discriminate between the generated potential outcome and the factual outcome. At this point, we have attained all potential outcomes for each unit under all tumor types. The training procedure optimizing the outcome generator and TF discriminator uses the minimax decision rule:
\begin{equation}
\begin{split}
    \text{min}_{g} \text{max}_{\phi} \ (\mathcal{L}_g -\beta \mathcal{L}_{\phi,g}),
    \label{Eqn4}
\end{split}
\end{equation}
where $\beta$ is a hyper-parameter controlling the trade-off between the outcome generator and discriminator. Compared to the deep regression task in the outcome generator, the TF discriminator is a relatively simple binary classification, which is easier to optimize. In every optimization iteration,  in order to get more accurate inferred potential outcomes to fool the discriminator based on the discriminator's current ability of telling which is factual outcome and which is counterfactual outcome, we can optimize $\text{min}_{g} \ ( \mathcal{L}_g -\beta \mathcal{L}_{\phi,g} )$ several times after we optimize $ \text{max}_{\phi} \ (-\beta \mathcal{L}_{\phi,g})$ one time.

\subsection{Targeted Group Analysis}

The proposed MTAL method can generate all potential outcomes for each unit across all tumor types, which can help basket trials increase sample size and thus increase statistical power, and remove the influence of heterogeneity among different tumor types. 

In basket trials, we must consider different configurations of effectiveness. For example, the drug may truly work for only one type of tumor due to the heterogeneity of tumors. Alternatively, it may actually work for all types of tumors, which means it works for the mutation regardless of the tumor types. Each of these configurations can lead to markedly different statistical properties~\citep{cunanan2017basket}. Therefore, we not only want to evaluate the treatment effect of the drug for the mutation (the whole population in the study) but also want to evaluate the effect of the drug for specific tumors (the sup-population in the study). In addition, most basket trials are conducted as single-arm trials without a control group and a primary endpoint is given by an objective response rate (ORP). We propose a new metric named targeted group response rate (TGOR) to better describe treatment effects in basket trials. TGOR describes the overall objective response rate for a given mutation or a given tumor type, which is defined as:
\begin{equation*}
 \text{TGOR}_{\text{mu}}  = \frac{1}{n\times k} \sum_{t=1}^k \sum_{i=1}^{n}y^{ti} 
\end{equation*}
and
\begin{equation*}
 \text{TGOR}_{\text{tu}}  = \frac{1}{n_c} \sum_{i=1}^{n_c}y^i,
\end{equation*}

where $n$ is the number of patients with that mutation and $n_c$ is the sub-sample who have that mutation and that specific cancer i.e., a subset of mutation sample $n$.

Our MTAL method can help remove heterogeneity across tumor types in basket trials when estimating the treatment effect for targeted mutation $\text{TGOR}_{\text{mu}}$, remove heterogeneity across patients with the same type of tumor when estimating the treatment effect for a targeted tumor $\text{TGOR}_{\text{tu}}$, and estimate the individualized treatment effects for an individual patient. Our method is also useful for studying rare cancers and cancers with rare genetic mutations by borrowing strength from more common cancers sharing the same mutation to infer the potential outcomes of existing patients under counterfactual cancer to increase sample size and statistical power.  

\section{Experiments and Analysis}

Because our method is the first model for estimating treatment effects for basket trials, no other baseline methods are available. To evaluate our model's estimation performance, we modify our model (by removing the deep feature selection module) to coordinate the settings in traditional treatment effect estimation (binary and multiple treatments) and use benchmarks (\textit{IHDP} and \textit{News}) to demonstrate our estimation performance on the counterfactual outcomes. We also use one synthetic basket trial dataset to demonstrate the method's stability in basket trials.

\subsection{Performance Evaluation on Estimating the Counterfactual Outcomes}

We coordinate our model to be compatible with the settings in traditional treatment effect estimation and conduct experiments on binary treatment benchmark \textit{IHDP} and multiple treatment benchmark \textit{News} with $2, 4, 8, \text{and} \ 16$ treatment options. 

\textbf{Datasets.} \textit{IHDP.} The IHDP data set is a commonly adopted benchmark collected by the Infant Health and Development Program~\citep{brooks1992ihdp}. These data are generated based on a randomized controlled trial where intensive high-quality care and specialist home visits were provided to low-birth-weight and premature infants. There are a total of 25 pre-treatment covariates and 747 units, including 608 control units and 139 treatment units. The outcome is the infants' cognitive test scores which can be simulated using the pre-treatment covariates and the treatment assignment information through the NPCI package \footnote{\url{https://github.com/vdorie/npci}}. In the IHDP data set, a biased subset of the treatment group is removed to simulate the selection bias ~\citep{shalit2017estimating}. We repeat these procedures 1000 times so as to conduct evaluations of the uncertainty of estimates. \textit{News.} The News data set was first introduced for binary treatments counterfactual estimation by ~\citep{johansson2016learning} and extended to multiple treatment benchmarks by ~\citep{schwab2018perfect}. The News benchmark includes 5000 randomly sampled news articles from the NY Times corpus and the opinions of a media consumer exposed to multiple news items. Each unit is a news item and the features are word counts. The outcome represents the reader’s opinion of the news item. The treatment options are various devices used for viewing news items, e.g. smartphones, tablets, desktops, televisions, or others. We use the extended multiple treatment data set according to the specification in ~\citep{schwab2018perfect}. A topic model is trained on the whole NY Times corpus to model consumers' preferences towards reading given news items on specific devices,  where $k + 1$ centroids are randomly picked in the topic space where $k$ represents the number of treatment options and the remaining is the control group. We use four different variants of this data set with 5000 units, 2870 features and $k = 2, 4, 8, \text{and} \ 16$ treatment options.

\textbf{Baselines.}
To evaluate the accuracy of our model's treatment effect estimation, we compare our multi-task adversarial learning net model with the following methods: k-nearest neighbor (kNN)~\citep{ho2007matching}, Causal forests (CF)~\citep{wager2018estimation},Random forests (RF)~\citep{breiman2001random}, Bayesian additive regression trees (BART)~\citep{chipman2010bart}, Generative adversarial nets for inference of ITE (GANITE)~\citep{yoon2018ganite}, Propensity score matching with logistic regression (PSM) ~\citep{ho2011matchit}, Treatment-agnostic representation network (TARNET)~\citep{shalit2017estimating}, Counterfactual regression network ($\text{CFRNET}_\text{wass}$)~\citep{shalit2017estimating}, local similarity preserved individual treatment effect estimation method (SITE)~\citep{yao2018representation},  and Perfect match (PM)~\citep{schwab2018perfect}.

\textbf{Parameter Settings.}
To ensure a fair comparison, we follow a standardized implementation \footnote{\url{https://github.com/d909b/perfect_match}} to realize hyperparameter optimization for IHDP and News data sets and extend the original binary treatment effect estimation methods to multiple treatments according to specifications in ~\citep{schwab2018perfect}. The hyperparameters of our method are chosen based on performance on the validation data set, and the searching range as shown in Table~\ref{hyperparameter}. MTAL is implemented using feed-forward neural networks with Dropout ~\citep{srivastava2014dropout} and the ReLU activation function. Adam~\citep{kingma2014adam} is adopted to optimize the objective function.

\begin{table}[t]
  \caption{Hyperparameters and ranges.}
  \label{hyperparameter}
  \centering
  \small
  \begin{tabular}{lll}
    \toprule
    \multicolumn{1}{c}{} & \multicolumn{1}{c}{IHDP}      & \multicolumn{1}{c}{News}      \\
    \midrule
    $\beta$   &  0, $\{10^{k}\}_{k=-6}^{2}$  & 0, $\{10^{k}\}_{k=-6}^{2}$\\ 
    \midrule
    $\lambda$,  $\alpha$ & 0, $\{10^{k}\}_{k=-6}^{-1}$ &  0, $\{10^{k}\}_{k=-6}^{-1}$\\
    \midrule
    No. layers  & 2, 3, 4, 5 &  2, 3, 4, 5\\
    \midrule
    Dim. layer & 50, 100, 150 & 50, 100, 150\\
    \midrule
    Batch size & 50$\times$2, 75$\times$2, 100$\times$2 &   30$k$, 40$k$, 50$k$\\
    \bottomrule
  \end{tabular}
\end{table}

\begin{table*}[t]

  \caption{Performance on IHDP and News data sets. We present mean $\pm$ standard deviation of $\sqrt{\epsilon_\text{PEHE}}$ and $\sqrt{\epsilon_\text{mPEHE}}$ on the test sets.} 
  \label{IHDPpehe}
  \centering
  \begin{tabular}{llllll}
    \toprule
    \multicolumn{1}{l}{} & \multicolumn{1}{l}{IHDP} & \multicolumn{1}{l}{News-2}  & \multicolumn{1}{l}{News-4} & \multicolumn{1}{l}{News-8}   & \multicolumn{1}{l}{News-16}                    \\
    \cmidrule(lr){2-6}  
    Method     & $\sqrt{\epsilon_\text{PEHE}}$    & $\sqrt{\epsilon_\text{PEHE}}$  & $\sqrt{\epsilon_\text{mPEHE}}$     & $\sqrt{\epsilon_\text{mPEHE}}$ & 
    $\sqrt{\epsilon_\text{mPEHE}}$   \\
    \midrule
    
    kNN     & $6.66\pm6.89$  & $18.14\pm1.64$   & $27.92\pm2.44$ & $26.20\pm2.18$  & $27.64\pm2.40$  \\
    PSM     & $2.70\pm3.85$  & $17.40\pm 1.30$   & $37.26\pm2.28$ & $30.50\pm1.70$  & $28.17\pm2.02$ \\
    RF    & $4.54\pm7.09$  & $17.39\pm1.24$   & $26.59\pm3.02$ & $23.77\pm2.14$  & $26.13\pm2.48$  \\
    CF    & $4.47\pm6.55$  & $17.59\pm 1.63$   & $23.86\pm 2.50$ & $22.56\pm2.32$ & $21.45\pm 2.23$  \\
    BART     & $2.57\pm3.97$  & $18.53\pm2.02$   & $26.41\pm 3.10$ & $25.78\pm2.66$   & $27.45\pm2.84$\\
    GANITE     & $5.79\pm8.35$  & $18.28\pm1.66$   & $24.50\pm2.27$ & $23.58\pm2.48$  & $25.12\pm3.53$  \\
    PD    & $5.14\pm6.55$  & $17.52\pm1.62$   & $20.88\pm3.24$ & $21.19\pm2.29$  & $22.28\pm2.25$ \\
    TARNET    & $1.32\pm 1.61$  & $17.17\pm1.25$   & $23.40\pm2.20$ & $22.39\pm2.32$  & $21.19\pm2.01$  \\
    CFRNET$_\text{wass}$     & $0.88\pm1.25$  & $ 16.93\pm1.12$   & $22.65\pm1.97$ & $21.64\pm1.82$  & $20.87\pm1.46$   \\
    PM & $ 0.84 \pm 0.61$  & $16.76 \pm 1.26$   & $21.58 \pm 2.58$  & $20.76\pm 1.86$  & $20.24\pm1.46$ \\
    SITE  & $\textbf{0.81}\pm \textbf{1.22}$  & $16.87\pm1.34$   & $22.33\pm2.08$ & $21.84\pm2.21$  & $20.88\pm1.52$ \\
    \midrule
    MTAL  & $1.06\pm1.28$  & $\textbf{16.58}\pm\textbf{1.20}$   & $\textbf{20.42}\pm\textbf{1.88}$ & $\textbf{19.98}\pm\textbf{2.01}$  & $\textbf{19.32}\pm\textbf{1.76}$\\
    \bottomrule
  \end{tabular}
\end{table*}

\begin{table*}[t]
  \caption{Performance on IHDP and News data sets. We present mean $\pm$ standard deviation of $\epsilon_\text{ATE}$  and $\epsilon_\text{mATE}$ on the test sets. }
   
  \label{IHDPate}
  \centering
  \begin{tabular}{llllll}
    \toprule
    \multicolumn{1}{l}{} & \multicolumn{1}{l}{IHDP} & \multicolumn{1}{l}{News-2}  & \multicolumn{1}{l}{News-4} & \multicolumn{1}{l}{News-8}   & \multicolumn{1}{l}{News-16}                    \\
    \cmidrule(lr){2-6}  
    Method     & $\epsilon_\text{ATE}$    & $\epsilon_\text{ATE}$ & $\epsilon_\text{mATE}$    & $\epsilon_\text{mATE}$ & 
    $\epsilon_\text{mATE}$   \\
    \midrule
    kNN     & $3.19\pm 1.49$  & $7.83\pm2.55$   & $19.40 \pm3.12$ & $15.11\pm2.34$  & $17.27\pm2.10$  \\
    PSM     & $0.49\pm 0.81$  & $4.89\pm 2.39$   & $30.19\pm2.47$ & $22.09\pm1.98$  & $18.81\pm1.74$ \\
    RF    & $0.64\pm 1.25$  & $5.50\pm 1.20$   & $18.03\pm3.18$ & $12.40\pm2.29$  & $15.91\pm2.00$  \\
    CF    & $0.65\pm1.24$  & $4.02\pm 1.33$   & $13.54\pm 2.48$ & $9.70\pm1.91$ & $8.37\pm 1.76$  \\
    BART     & $0.53\pm1.02$  & $5.40\pm1.53$   & $17.14\pm 3.51$ & $14.80\pm2.56$   & $17.50\pm2.49$\\
    GANITE     & $0.98\pm1.90$  & $4.65\pm 2.12$   & $13.84\pm2.69$ & $11.20\pm 2.84$  & $13.20\pm3.28$  \\
    PD    & $1.37\pm1.65$  & $4.69\pm3.17$   & $8.47\pm4.51$ & $7.29\pm2.97$  & $10.65\pm2.22$ \\
    TARNET    & $ 0.24\pm 0.29$  & $4.58\pm1.29$   & $13.63\pm2.18$ & $9.38\pm 1.92$  & $8.30\pm1.66$  \\
    CFRNET$_\text{wass}$     & $0.20\pm0.24$  & $ 4.54\pm1.48$   & $12.96\pm 1.69$ & $8.79\pm 1.68$  & $8.05\pm1.40$   \\
    PM & $  0.24 \pm 0.20$  & $3.99 \pm 1.01$   & $10.04 \pm 2.71$  & $6.51\pm 1.66$  & $ 5.76\pm1.33$ \\
    SITE  & $\textbf{0.18}\pm\textbf{0.23}$  & $4.53\pm1.32$   & $12.75\pm1.88$ & $9.01\pm1.86$  & $8.63\pm1.41$ \\
    \midrule
    MTAL  & $0.34\pm0.28$  & $\textbf{3.88}\pm\textbf{1.11}$   & $\textbf{8.01}\pm\textbf{1.43}$ & $\textbf{5.97}\pm\textbf{1.58}$  & $\textbf{5.12}\pm\textbf{1.31}$\\
    \bottomrule
  \end{tabular}
\end{table*}

\begin{figure*}[t]
    \centering
    \includegraphics[width=1\textwidth]{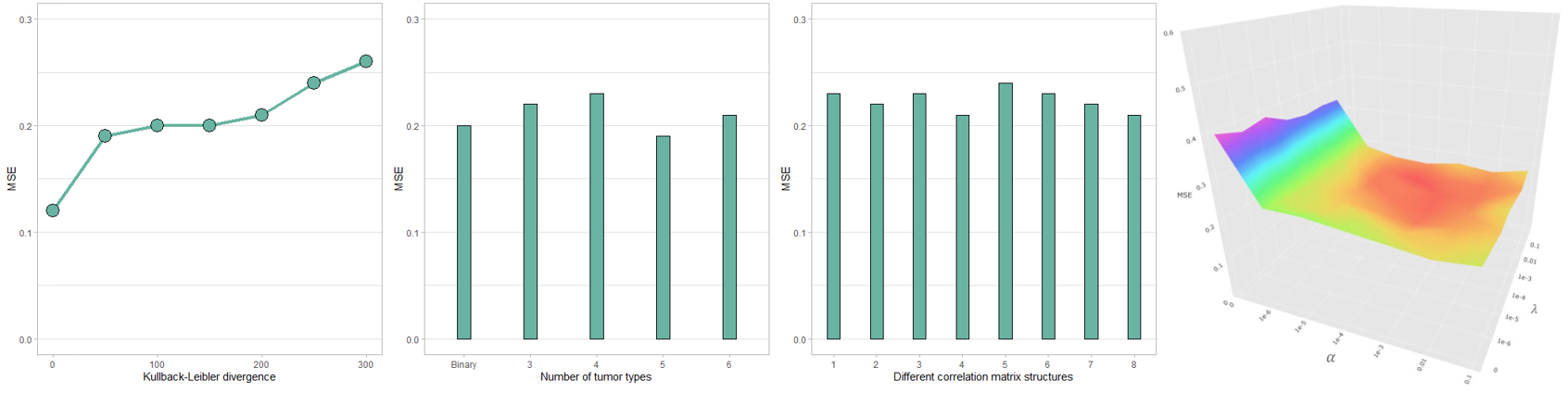}
    \caption{The results for synthetic basket trial data sets.}
    \label{fig: syntheticdata}
\end{figure*}

\begin{figure*}[t]
    \centering
    \includegraphics[width=0.95\textwidth]{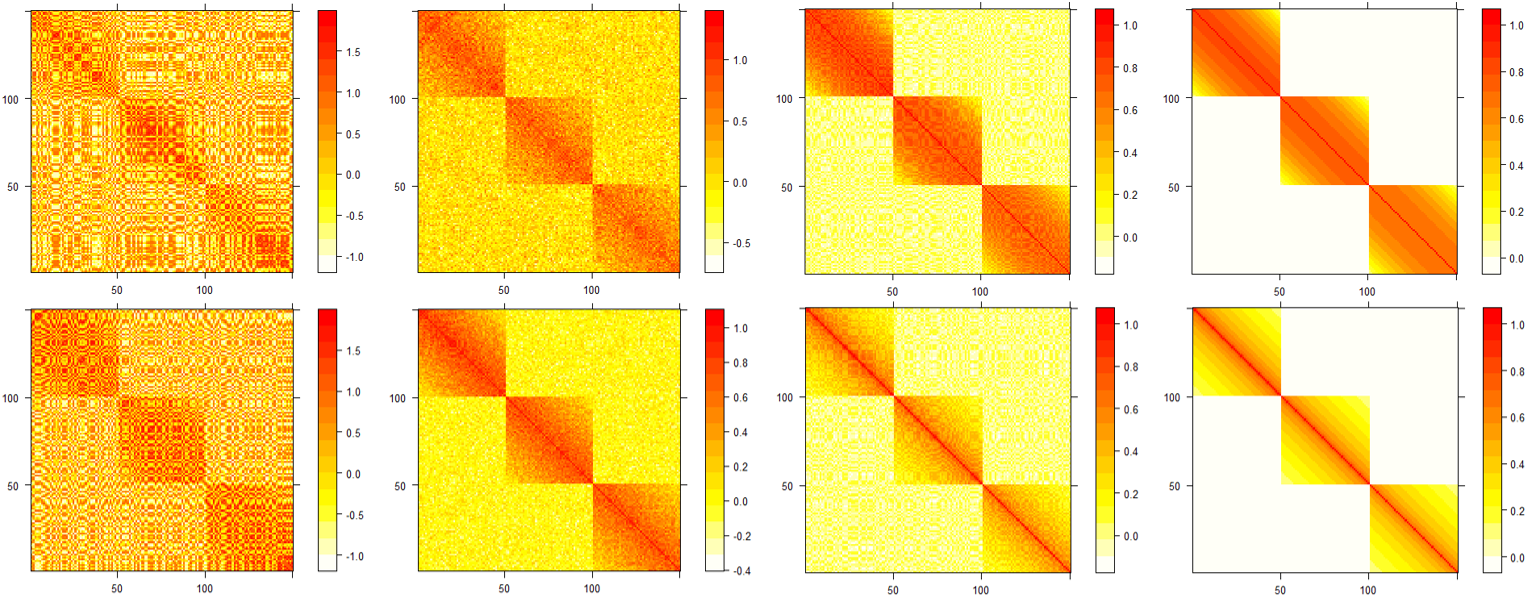}
    \caption{Different covariates correlation structures.}
    \label{fig: correlation}
\end{figure*}

\textbf{Results and Analysis.}
We adopt two commonly used evaluation metrics. The first one is the error in average treatment effect (ATE) estimation defined as $\epsilon_\text{ATE}  = |\text{ATE} - \widehat{\text{ATE}}|$, where $\text{ATE}=\frac{1}{n}\sum_{i=1}^{n}(Y_1^i - Y_0^i)$ and $\widehat{\text{ATE}}$ is an estimated \text{ATE}. The second one is the error of expected precision in estimation of heterogeneous effect (PEHE)~\citep{hill2011bayesian}, which is defined as $\epsilon_\text{PEHE}  = \frac{1}{n}\sum_{i=1}^{n}(\text{ITE}_i-\widehat{\text{ITE}}_i)^2$, where $ \text{ITE}_i = Y_1^i - Y_0^i$ and $\widehat{\text{ITE}}_i$ is an estimated \text{ITE} for unit $i$. In addition, for multiple treatment evaluations, we follow the definitions in ~\citep{schwab2018perfect}, where both $\epsilon_\text{PEHE}$  and $\epsilon_\text{ATE}$ can be extended to multiple treatments by averaging PEHE and ATE between every possible pair of treatments. They are defined as $\epsilon_\text{mPEHE}  =\frac{1}{\binom k2}\sum_{t=1}^{k}\sum_{j=1}^{t}\epsilon_{\text{PEHE},t,j}$ and $\epsilon_\text{mATE}  = \frac{1}{\binom k2}\sum_{t=1}^{k}\sum_{j=1}^{t}\epsilon_{\text{ATE},t,j}$. Table~\ref{IHDPpehe} and table ~\ref{IHDPate} show the performance of our method and baseline methods on the IHDP and News data sets. MTAL achieves the best performance with respect to PEHE and ATE for News data sets with different numbers of treatment options. For the IHDP data set, MTAL still has competitive performance when compared to the best baseline methods with respect to PEHE and ATE. The results on these two benchmarks for conventional binary and multiple treatments effects estimation can demonstrate that our method is capable of precisely estimating the treatment effects.

\subsection{Synthetic Basket Trial Data Set}
\label{simulated_data}
\textbf{Dataset.} 
To evaluate our model's performance for basket trials, we simulate one synthetic data set which mimics the characteristics of a basket trial. Because different types of tumors may have different predictor variables, which may be a subset of all observed covariates, we use different subsets of the observable covariates to generate the outcomes for different tumor types. To mimic the real situation further, we consider different covariance matrices in the covariates simulation. For example, the covariates predicting outcomes in each tumor type are taken to have stronger correlations than covariates predicting outcomes for other tumor types. 

We generate a set of synthetic data sets which reflects the complexity of observational medical records data. The sample size for tumor type $k$ is $n_k$, where $k=1,2,...,K$. Therefore, the total sample size is $n = \sum_{k=1}^{K} n_k$ units. The predictor variables for tumor type $k$ are $x_k \in \mathbb{R}^{d}$. The potential outcomes $y_k$ for tumor type $k$ are generated as $y_k|x_k \sim \cos{({(w_k^\intercal x_k)}^2)}$, where $w_k \sim Uniform((-1,1)^{d \times 1})$. The vector of all observed covariates $x = ( x_1^\intercal, x_2^\intercal,..., x_K^\intercal)^\intercal$ is sampled from a multivariate normal distribution with mean $\mu_k$ and different random positive definite covariance matrices $\Sigma$. By varying the value of $\mu_k$, data with different levels of selection bias are generated~\citep{yoon2018ganite, yao2018representation}. Let $D$ be the diagonal matrix with the square roots of the diagonal entries of  $\Sigma$ on its diagonal, i.e., $D = \sqrt{diag(\sigma)}$, then the correlation matrix is given by $R= D^{-1}\Sigma D^{-1} $. We simulate correlation matrix to better explain the relationship of covariates among and within different tumor types, instead of directly simulating covariates matrix. We use algorithm 3 in ~\citep{hardin2013method} to simulate positive definite correlation matrices consisting of different within tumor type correlations and between tumor type correlations. Our correlation matrices are based on the hub correlation structure which has a known correlation between a hub variable and each of remaining variables~\citep{zhang2005general, langfelder2008defining}. Each variable in a tumor type is correlated with the hub-variable with decreasing strength from specified maximum correlation to minimum correlation and different tumor types are generated independently or with weaker correlation among tumor types. Defining the first variable as the hub, for the $i$th variable $(i = 2, 3, . . . , d)$, the correlation between it and the hub-variable in one tumor type is given by $R_{i,1} = \rho_{\text{max}} - \left ( \frac{i-2}{d-2} \right )^{\gamma} (\rho_{\text{max}}-\rho_{\text{min}})$, where $ \rho_{\text{max}} $ and $ \rho_{\text{min}} $ are specified maximum and minimum correlations, and the rate $\gamma$ controls rate at which correlations decay. 

After specifying the relationship between the hub variable and the remaining variables in one tumor type, we use the Toeplitz structure to fill out the remainder of the hub correlation matrix and get the hub-Toeplitz correlation matrix $R_k$ for tumor type $k$. Here, $R$ is the $d \times d$ matrix having the blocks $R_1, R_2,..., R_K$ along the diagonal and zeros at off-diagonal elements. This yields a correlation matrix with nonzero correlations within tumor types and zero correlation among tumor types. The amount of correlation among tumor types that can be added to the positive-definite correlation matrix $R$ is determined by its smallest eigenvalue. 

\textbf{Results and Analysis.}
The mean squared error is used as the performance metric to evaluate our model under the  settings of binary or multiple tumor types,  different selection bias, and different correlation matrix for observed covariates. The mean squared error is defined as $\text{MSE}=\frac{1}{ n \times K} \sum_{i=1}^{n}\sum_{k=1}^{K}\left(y_k(x_i)-\hat{y}_k(x_i)\right)^2$, where $y_k(x_i)$ and $\hat{y}_k(x_i)$ are factual and estimated outcomes for unit $i$ with features $x_i$ and tumor type $k$, respectively.

We simulate 5 data sets with $2, 3, 4, 5 ,\text{and} \ 6$ tumor types, separately. From the second figure in Fig.~\ref{fig: syntheticdata}, our MTAL performs relatively steadily for binary and multiple tumor types. To evaluate the performance with respect to selection bias, Kullback-Leibler divergence is adopted to quantify selection bias among different tumor types. Here, we use the data sets with binary tumor types. All observed covariates in two tumor types are generated by a multivariate normal distribution with mean $0$ and different mean $\mu_1$ for the remaining tumor types, so different values of $\mu_1$ represent different Kullback-Leibler divergences; i.e., selection bias between two tumor types. From the first figure in Fig.~\ref{fig: syntheticdata}, for the MTAL method, MSE increases modestly with increasing selection bias. To evaluate the sensitivity of the MTAL method to the correlation structure of the covariates, we generate 8 different correlation matrices with different levels of correlation for variables within each tumor type and among different tumor types in Fig.~\ref{fig: correlation}. From the third figure in Fig.~\ref{fig: syntheticdata}, we find that the MSE in our MTAL method is not sensitive to the structure of the correlation matrices. In addition, from the fourth figure in Fig.~\ref{fig: syntheticdata}, the performance of our model, with respect to MSE, is significantly improved compared to the models without $L_1$ or $L_2$ penalties. Also, the overall performance on different combinations of hyperparameters of $L_1$ and $L_2$ penalties is stable over a large range of tuning parameter values, which confirms the effectiveness and robustness of deep feature selection in our MTAL method.

\section{Conclusion}
In this paper, we propose a multi-task adversarial learning (MTAL) method by incorporating feature selection multi-task deep learning and adversarial learning to remove heterogeneity of tumor types in basket trials. To the best of our knowledge, our model is the first work introducing machine learning and causal inference to the task of analyzing basket trial data. It not only improves the basket trial analysis but also has its superiority over state-of-the-art methods in estimating multiple treatment effects for observational data. In future work, we will follow this direction to apply causal inference models and machine learning methods into more medical practical applications, such as umbrella, platform trials, and so on.

% \newpage
\section*{Institutional Review Board (IRB)}
Our research does not require IRB approval.
% \bibliography{jmlr-sample}

\end{document}